\documentclass{article} \usepackage{preprint,times}

\usepackage{amsmath,amsfonts,bm}

\def\eqref#1{equation~\ref{#1}}

\def\1{\bm{1}}

\DeclareMathAlphabet{\mathsfit}{\encodingdefault}{\sfdefault}{m}{sl}
\SetMathAlphabet{\mathsfit}{bold}{\encodingdefault}{\sfdefault}{bx}{n}

\usepackage{hyperref}
\usepackage{url}
\usepackage{graphicx}
\usepackage{adjustbox}
\usepackage[table]{xcolor}

\definecolor{bestblue}{HTML}{B8D1EA}
\definecolor{secondblue}{HTML}{EFF5FB}
\newcommand{\bestresult}[1]{\cellcolor{bestblue}\textbf{#1}}
\newcommand{\secondresult}[1]{\cellcolor{secondblue}\underline{#1}}
\usepackage{booktabs}
\usepackage{multirow}
\usepackage{amsmath}
\usepackage{amssymb}
\usepackage{subcaption}
\usepackage{pifont}
\usepackage{float}

\newcommand{\papertablestyle}{\small\setlength{\tabcolsep}{5pt}\renewcommand{\arraystretch}{1.08}}

\graphicspath{{Figure/}}

\title{Asymmetric Looped Transformers: \\ Fixed Global Representations for an Iterative Decoder}

\author{\begin{tabular}{c}
Cheng Liang\textsuperscript{1},\quad
Tengfei Zhang\textsuperscript{2},\quad
Wenjie DU\textsuperscript{3},\quad
Jianyu Wu\textsuperscript{1},\quad
Changlian Ma\textsuperscript{4} \\[4pt]
Shuchen Pu\textsuperscript{2},\quad
Ya Zhang\textsuperscript{1},\quad
Yanfeng Wang\textsuperscript{1},\quad
Zhouhan Lin\textsuperscript{1,*}
\end{tabular}
\par\vspace{7pt}
{\small
\textsuperscript{1} Shanghai Jiaotong University \\
\textsuperscript{2} University of Science and Technology of China \\
\textsuperscript{3} Nanyang Technological University\quad
\textsuperscript{4} Nanjing University \\[3pt]
\textsuperscript{*} Corresponding author
\par}
}

\hypersetup{
  pdftitle={Asymmetric Looped Transformers: Fixed Global Representations for an Iterative Decoder},
  pdfauthor={Cheng Liang, Tengfei Zhang, Wenjie DU, Jianyu Wu, Changlian Ma, Shuchen Pu, Ya Zhang, Yanfeng Wang, Zhouhan Lin},
  pdfsubject={AsymLoop},
  pdfkeywords={looped transformers, recurrent language models}
}

\begin{document}
\raggedbottom

\maketitle

\begin{abstract}
Looped Transformers increase effective depth by repeatedly applying shared layers, offering a way to scale computation without increasing parameter count. However, full-stack looping places context construction and iterative state refinement within the same recurrent computation, without explicitly distinguishing their roles. We introduce the \textbf{Asymmetric Looped Transformer (AsymLoop)}, which separates single-pass context encoding from recurrent state refinement. A single-pass causal encoder produces layer-specific Global key--value (KV) caches that remain fixed across iterations, while a weight-tied recurrent decoder generates Local KV from its evolving state. Joint attention over both sources under a single softmax allows each update to integrate persistent contextual information with the current intermediate state. For training, we apply a heavy-tail-guided layerwise learning-rate method to the encoder and decoder. Experiments on FineWeb-Edu across four model scales, from 60M to 1B, show consistently lower validation perplexity and higher average downstream accuracy than Full Loop under matched effective depth and training token budgets. At 1B, AsymLoop improves average zero-shot accuracy across seven commonsense reasoning benchmarks from 49.58\% to 50.85\%. Further experiments extend these benefits to longer training budgets, an OLMo-style backbone, and sparse mixture-of-experts models. These results support asymmetric context encoding and state refinement as an effective design for recurrent language models.
\end{abstract}

\section{Introduction}
\label{sec:intro}

Scaling language models typically involves increasing parameters, training data, and computation together~\citep{henighan2023scaling}. Looped Transformers offer another dimension for scaling computation: repeatedly applying a shared block increases effective depth without assigning independent parameters to every executed layer~\citep{dehghani2019universal,lan2020albert}. Recent work further demonstrates that recurrent depth can improve reasoning performance through additional computation at test time~\citep{geiping2025recurrentdepth}. This opportunity makes the organization of recurrence a central design concern. Performance depends not only on how many iterations a model executes, but also on which computations it repeats and how contextual information remains available across iterations.

Recurrence can be organized through different layer topologies, depending on which layers are repeated and which non-recurrent layers precede or follow them~\citep{huang2026loopedmodels}. A straightforward choice is to loop the entire Transformer stack, feeding the output of each pass back into the same stack for the next~\citep{zhu2025scalinglatent}; we refer to this topology as \emph{Full Loop}. However, analyses of pretrained language models suggest that different depths support distinct computational roles, including contextualizing input tokens, refining features, and forming predictions~\citep{lad2025remarkablerobustnessllmsstages}. Repeating the entire stack places these roles together within the recurrent computation, without explicitly separating input processing from iterative refinement. Analyses of looped models likewise observe multiple feedforward-like processing stages recurring within each iteration~\citep{blayney2026mechanisticanalysisloopedreasoning}. These findings motivate organizing recurrence around a division of roles: distinguishing context construction that can be performed once from state refinement to which additional computation can be allocated.

Partial looping has been explored in recurrent language models, repeating only a selected subset of layers while executing the remaining layers once~\citep{geiping2025recurrentdepth,huang2026loopedmodels}. Controlled comparisons with matched parameter counts, effective depth, and training budgets report improvements over full-stack looping on several multi-step reasoning benchmarks~\citep{huang2026loopedmodels}. Such designs make the distinction between recurrent and non-recurrent computation explicit, but choosing the scope of recurrence alone does not determine how the two components should work together. We argue that their different computational roles should guide a broader \emph{asymmetric design}: non-recurrent layers can specialize in context construction, while recurrent layers perform iterative state updates, with information exchange connecting these two types of computation.

We introduce the \textbf{Asymmetric Looped Transformer (AsymLoop)}, which connects a single-pass causal encoder to a weight-tied recurrent decoder through joint attention to fixed Global KV and evolving Local KV. A causal encoder runs once and projects its output into per-decoder-layer Global key--value (KV) caches. These caches serve as contextual anchors that remain fixed across all iterations of a forward pass. A weight-tied decoder generates Local KV from its current hidden states at each iteration. The two sources offer complementary views: Global KV preserves access to the context encoded by the encoder, while Local KV reflects the decoder's current intermediate state. Queries allocate attention jointly across both sources under a single softmax, enabling each update to integrate global contextual information with the current local state. All components are trained end to end.

For optimization, we adopt the heavy-tail-guided layerwise learning-rate method of \citet{he2026one} for the encoder and decoder. The same spectral allocation rule is applied to matrices in both components. Encoder--decoder learning-rate differences arise from their measured spectral statistics. We use Asymmetric Learning Rate Schedule (ALRS) as a shorthand for this training configuration and examine the resulting allocation patterns and their empirical effects in our experiments.

Our main contributions are:
\begin{itemize}
    \item \textbf{An asymmetric recurrent architecture.} We jointly design one-time context encoding and weight-tied recurrent updates, connecting them through joint attention to fixed Global KV and dynamic Local KV.
    \item \textbf{Analysis of context access under recurrence.} Through component ablations and comparisons across recurrent depths, we examine the roles of fixed Global KV and evolving Local KV in recurrent state refinement.
    \item \textbf{Performance gains across model sizes.} Across four model sizes, AsymLoop achieves lower validation perplexity and higher average zero-shot accuracy than Full Loop under matched training budgets and effective depth. Perplexity improves by 0.98 and 0.30 at 135M and 350M, respectively, while the seven-task average accuracy at 1B rises from 49.58\% to 50.85\%.
\end{itemize}

\section{Related Work}
\label{sec:related}

\textbf{Looped Transformers and Recurrent Depth.} 
Weight-tied recurrence in Transformers traces back to Universal Transformers~\citep{dehghani2019universal} and ALBERT~\citep{lan2020albert}, which established that parameter sharing across layers can achieve competitive linguistic representations. Theoretical investigations have formalized looped Transformers as programmable computers~\citep{giannou2023looped} capable of learning complex in-context algorithms~\citep{yang2023looped} and latent thoughts~\citep{saunshi2025reasoning}. More recently, scaling up test-time compute through recurrent depth has emerged as a promising alternative to chain-of-thought token generation~\citep{geiping2025recurrentdepth}. Studies such as Parcae~\citep{parcae} and FLT~\citep{flt} explore stability in looped models. Beyond full-stack recurrence, partial-loop architectures already separate recurrent computation from non-recurrent input and output processing~\citep{geiping2025recurrentdepth}. Our focus is the information exchange between these components: AsymLoop provides the recurrent decoder with layer-specific Global KV from a single-pass encoder, jointly accessed with evolving Local KV at every iteration.

\textbf{Causal Encoder-Decoder Architectures.} 
Standard autoregressive language modeling typically employs decoder-only stacks. YOCO lets upper layers reuse KV produced by lower causal layers~\citep{yoco}. The Causal Encoder-Decoder (CED) proposed in DeepSeek-V4.1-Flash~\citep{deepseekai2026deepseekv41flashpushinglimitskv} divides the Transformer into an encoder that generates a static Global KV cache once, and a decoder that attends to this cache alongside standard causal dependencies. While CED demonstrated KV cache savings during inference, it was conceived as a non-recurrent ($K=1$) feedforward model. Their asymmetry also suggests a way to organize recurrent language models: build a causal representation of the input once, then let a recurrent stack iteratively refine its state while retaining access to that representation.

\textbf{Heavy-Tailed Self-Regularization and Layerwise Optimization.} 
Heavy-Tailed Self-Regularization (HT-SR) characterizes weight spectra through fitted power-law tails~\citep{martin2019traditional,martin2021predicting}. \citet{he2026one} use these spectral exponents to allocate learning rates across layers and modules. We apply their allocation rule to the non-recurrent encoder and the weight-tied decoder. Encoder parameters are used once per forward pass, whereas decoder parameters are reused across iterations and accumulate gradient contributions from each use. This difference motivates allowing their learning rates to vary.

\section{The Asymmetric Looped Architecture}
\label{sec:method}

\begin{figure}[H]
    \centering
    \includegraphics[width=\linewidth]{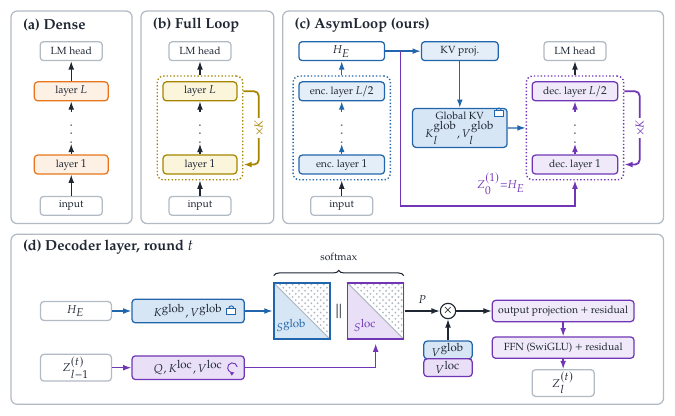}
    \caption{\textbf{AsymLoop architecture and Global--Local attention.}
    \textbf{(a)} Dense executes each layer once.
    \textbf{(b)} Full Loop repeats the entire stack with shared parameters.
    \textbf{(c)} AsymLoop runs a causal encoder once and a shared-parameter decoder for $K$ iterations. The encoder output provides layer-specific Global KV that remains fixed across iterations.
    \textbf{(d)} Each decoder layer jointly attends to fixed Global KV and Local KV computed from its current state, using a single softmax with causal masking for both sources.}
    \label{fig:arch}
\end{figure}

\subsection{Asymmetric Encoder--Decoder Design}
\label{subsec:method-topology}

As shown in Figure~\ref{fig:arch}, AsymLoop consists of a non-recurrent causal encoder and a weight-tied recurrent decoder. The encoder uses the same causal Transformer blocks as a standard decoder-only language model, without changing their architecture. We split $L$ physical layers into $E$ encoder layers and $D$ decoder layers, with $E=D=L/2$ in our experiments. A single forward pass through the encoder produces contextual representations $H_E\in\mathbb{R}^{N\times d}$, where $N$ is the sequence length and $d$ is the hidden dimension. The encoder output serves as the input to the first decoder iteration. Separate projection matrices for each decoder layer map this output into layer-specific Global KV caches, which are reused across all $K$ iterations.

The decoder runs for $K$ iterations, feeding the output of each iteration into the next. Denote its $D$ layers by $\mathcal{D}_1,\ldots,\mathcal{D}_D$, and the Global KV for layer $\mathcal{D}_l$ by $\mathcal{G}_l$, defined below. At iteration $t$, $X^{(t)}$ denotes the input to the entire decoder, and $Z_l^{(t)}$ denotes the output of decoder layer $\mathcal{D}_l$. Thus, $Z_D^{(t)}$ is the output of the final decoder layer $\mathcal{D}_D$ and of the entire decoder at that iteration. The first iteration receives the encoder output; each subsequent iteration receives the preceding iteration's decoder output. Within each iteration, the first layer processes $X^{(t)}$, and the remaining layers process the preceding layer's output:
\begin{equation}
    \begin{aligned}
        X^{(1)}&=H_E,\qquad X^{(t)}=Z_D^{(t-1)} && (t=2,\ldots,K),\\
        Z_1^{(t)}&=\mathcal{D}_1\!\left(X^{(t)};\mathcal{G}_1\right),\\
        Z_l^{(t)}&=\mathcal{D}_l\!\left(Z_{l-1}^{(t)};\mathcal{G}_l\right) && (l=2,\ldots,D).
    \end{aligned}
    \label{eq:asym-recurrence}
\end{equation}
Each $\mathcal{D}_l$ has distinct parameters shared across all $K$ iterations. The final state $Z_D^{(K)}$ is used for next-token prediction, and the model is trained end to end through all iterations. 

\subsection{Global--Local Fusion via Joint Attention}
\label{subsec:method-attention}

Each decoder layer attends to two sources: fixed Global KV from the encoder and evolving Local KV from the current decoder state. For layer $l$, Global KV is obtained using layer-specific projections:
\begin{equation}
    \bar K_l^{\mathrm{glob}}=H_EW_l^{K,\mathrm{glob}},\qquad
    V_l^{\mathrm{glob}}=H_EW_l^{V,\mathrm{glob}}.
    \label{eq:asym-global-kv}
\end{equation}

At each iteration, the layer projects its current RMS-normalized input into queries and Local KV. Before attention, we apply QKNorm~\citep{henry2020query}, followed by RoPE, to queries and both Global and Local keys. The resulting Global keys and values form $\mathcal{G}_l=(K_l^{\mathrm{glob}},V_l^{\mathrm{glob}})$, which remains fixed across all $K$ iterations of a forward pass; queries and Local KV are recomputed at each iteration.

Each attention head reads both KV sources under a single softmax. Omitting layer and iteration indices, the attention output is
\begin{equation}
    A=\operatorname{softmax}\!\left(
    \frac{[Q(K^{\mathrm{glob}})^\top\mid Q(K^{\mathrm{loc}})^\top]}{\sqrt{d_h}}
    +[M\mid M]\right)
    \begin{bmatrix}V^{\mathrm{glob}}\\V^{\mathrm{loc}}\end{bmatrix},
    \label{eq:asym-joint-attention}
\end{equation}
where $d_h$ is the head dimension and $M$ is the causal mask. The two sources are concatenated along the key-position axis; a query at position $i$ can access only positions $j\leq i$ in either source. The shared softmax allocates attention between encoder context and the current decoder state. Standard output projection, residual connections, and FFN follow.

\subsection{Application of Spectral Layerwise Learning Rates}
\label{subsec:method-alrs}

We use the layerwise learning-rate method of \citet{he2026one} to train AsymLoop. The method uses spectral exponents from Heavy-Tailed Self-Regularization (HT-SR)~\citep{htsr} to assign learning rates to individual weight matrices in the encoder and decoder under the same allocation rule.

For a weight matrix $W_m$, let $\mathcal{T}_m$ contain the largest half of its squared singular values and let $\lambda_{m,\min}$ be the smallest value in this tail. At refresh step $s$, we estimate the exponent of the power-law density $p(\lambda)\propto\lambda^{-\alpha}$ using
\begin{equation}
    \alpha_m(s)=1+\frac{|\mathcal{T}_m|}
    {\displaystyle\sum_{\lambda\in\mathcal{T}_m}\log(\lambda/\lambda_{m,\min})}.
    \label{eq:asym-hill}
\end{equation}
Following the allocation rule of \citet{he2026one}, larger exponents receive larger learning-rate multipliers. With $\alpha_{\min}(s)$ and $\alpha_{\max}(s)$ denoting the extrema across the attention, FFN, and Global KV projection matrices, the target learning rate is
\begin{equation}
    \eta_m^{\mathrm{target}}(u;s)=\eta_{\mathrm{base}}(u)
    \left[1+4\frac{\alpha_m(s)-\alpha_{\min}(s)}
    {\alpha_{\max}(s)-\alpha_{\min}(s)}\right],
    \label{eq:asym-alrs}
\end{equation}
where $\eta_{\mathrm{base}}(u)$ is the common warmup/cosine schedule at training progress $u$. This maps the exponents to multipliers in $[1,5]$. Following \citet{he2026one}, the input embedding and output head use the upper multiplier; normalization parameters retain the base schedule.

Following \citet{he2026one}, ALRS periodically updates the learning-rate multipliers during the first 20\% of training, then holds them fixed while the base learning-rate schedule continues. Scheduling details are provided in Appendix~\ref{app:alrs-schedule}.

\section{Experiments}
\label{sec:experiments}

We evaluate AsymLoop on LLaMA-style architectures at the 60M, 135M, 350M, and 1B scales following~\citet{he2026one}. Section~\ref{subsec:exp_setup} defines the baselines, training budgets, and effective-depth comparisons, with full configurations and hardware details in Appendix~\ref{app:reproducibility-details}. We then report pre-training results across model sizes (Section~\ref{subsec:main_scaling}), downstream performance (Section~\ref{subsec:downstream}), the effect of recurrent depth (Section~\ref{subsec:recurrence-validation}), ablation studies (Section~\ref{subsec:ablations}), and transfer to an OLMo-style backbone (Appendix~\ref{app:olmo-transfer}).

\subsection{Experimental Setup}
\label{subsec:exp_setup}

\textbf{Pre-training.} All model families in the main experiments use the LLaMA-style backbone configurations following~\citet{he2026one} and are trained from scratch on FineWeb-Edu~\citep{fineweb}. To study the scaling behavior of AsymLoop, we evaluate the 60M, 135M, 350M, and 1B configurations. Training token budgets follow the Chinchilla scaling law~\citep{hoffmann2022training}, which prescribes approximately $20$ training tokens per parameter, giving 1.68B, 3B, 7B, and 20B tokens at the four model sizes. Detailed data splits, model configurations, and training budgets are provided in Appendix~\ref{app:reproducibility-details}.

\textbf{Baselines and depth matching.} We compare AsymLoop with two vanilla baselines: a standard non-recurrent Transformer (\emph{Dense}), and a homogeneous looped Transformer that repeats the entire Transformer stack with weights shared across iterations (\emph{Full Loop}). We compare recurrent models at matched effective depth, defined as the total number of Transformer layers executed per forward pass. 
The additional parameters and computation introduced by Global KV are detailed in Appendix~\ref{app:computational-cost}.

\begin{figure}[H]
\centering
\includegraphics[width=0.98\linewidth]{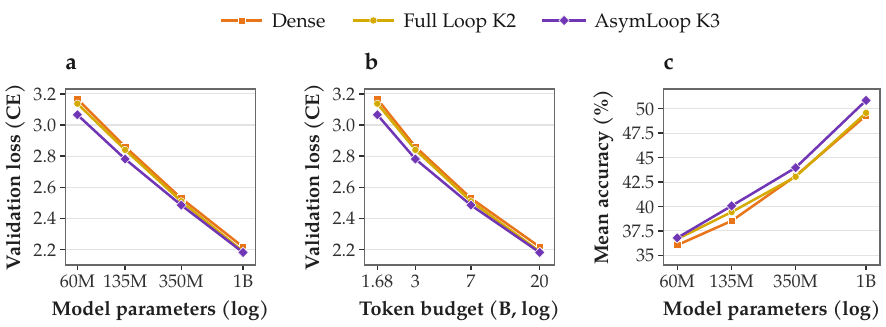}
\caption{\textbf{Validation and zero-shot performance across model sizes.} Full Loop ($K=2$) and AsymLoop ($K=3$) have the same effective depth. \textbf{(a, b)} Validation cross-entropy versus model parameter count and training token budget, respectively. \textbf{(c)} Average zero-shot accuracy across seven tasks.}
\label{fig:scaling_benchmark}
\end{figure}

\textbf{Evaluation.} We report validation cross-entropy and perplexity on held-out FineWeb-Edu data to assess language modeling performance. We further evaluate the pretrained models on seven commonsense reasoning benchmarks under zero-shot and five-shot settings: PIQA~\citep{bisk2020piqa}, Social IQa~\citep{sap2019socialiqa}, HellaSwag~\citep{zellers2019hellaswag}, WinoGrande~\citep{sakaguchi2021winogrande}, ARC-Challenge (ARC-c)~\citep{clark2018think}, ARC-Easy (ARC-e)~\citep{clark2018think}, and OpenBookQA (OBQA)~\citep{mihaylov2018can}. We use the standard prompts from the LM Evaluation Harness~\citep{eval-harness} and report per-task accuracy and the unweighted mean across tasks. Evaluation settings are summarized in Appendix~\ref{app:reproducibility-details}.

\subsection{Pre-training under the Chinchilla Scaling Law}
\label{subsec:main_scaling}

To examine the scalability of AsymLoop, we jointly increase model size and training token budget following the Chinchilla scaling law. Figure~\ref{fig:scaling_benchmark}(a, b) shows that AsymLoop consistently achieves lower validation loss than Dense and Full Loop across all four scales, from 60M to 1B. Its advantage over Full Loop is maintained at matched effective depth and training budgets within each scale. These results demonstrate the effectiveness of AsymLoop across model sizes, with sustained improvements at the billion-parameter scale. These gains also extend to an OLMo-style backbone~\citep{groeneveld-etal-2024-olmo}, where AsymLoop reduces validation perplexity by 14.8\% relative to the depth-matched Full Loop (Appendix~\ref{app:olmo-transfer}).

AsymLoop consistently improves both language-modeling loss and zero-shot downstream accuracy over Full Loop across all evaluated scales. Notably, the downstream advantage grows monotonically with model scale, from 0.13 percentage points at 60M to 1.27 points at 1B. We examine the task-level results in Section~\ref{subsec:downstream}.

\begin{table}[H]
\caption{\textbf{Zero- and five-shot commonsense reasoning accuracy ($\uparrow$, $\%$).} Full Loop ($K=2$) and AsymLoop ($K=3$) have the same effective depth. Avg. is the mean across seven tasks. Darker/lighter blue mark the best/second-best results at each model size and shot setting.}
\label{tab:downstream}
\centering
\papertablestyle
\setlength{\tabcolsep}{3pt}
\renewcommand{\arraystretch}{0.94}
\setlength{\aboverulesep}{1.2pt}
\setlength{\belowrulesep}{1.2pt}
\begin{tabular}{clrrrrrrrr}
\toprule
\textbf{Scale} & \textbf{Model} & \textbf{OBQA} & \textbf{WinoGrande} & \textbf{ARC-c} & \textbf{ARC-e} & \textbf{HellaSwag} & \textbf{SIQA} & \textbf{PIQA} & \textbf{Avg.} \\
\midrule
\multicolumn{10}{c}{\textbf{Zero-shot}} \\
\cmidrule(lr){1-10}
\multirow{3}{*}{\textbf{60M}}
 & Dense & \bestresult{17.80} & \bestresult{50.83} & 17.41 & 45.33 & \secondresult{27.39} & 34.54 & \secondresult{59.03} & 36.05 \\
 & Full Loop & \secondresult{16.80} & 50.28 & \bestresult{20.22} & \secondresult{46.51} & 27.14 & \bestresult{37.00} & 58.71 & \secondresult{36.66} \\
 & \textbf{AsymLoop} & \bestresult{17.80} & \secondresult{50.59} & \secondresult{18.17} & \bestresult{47.31} & \bestresult{27.72} & \secondresult{36.49} & \bestresult{59.41} & \bestresult{36.79} \\
\cmidrule(lr){3-10}
\multirow{3}{*}{\textbf{135M}}
 & Dense & 18.80 & 48.54 & 21.42 & \secondresult{52.90} & 29.06 & 37.41 & 61.53 & 38.52 \\
 & Full Loop & \secondresult{19.80} & \bestresult{51.85} & \bestresult{22.87} & 51.52 & \secondresult{29.09} & \bestresult{38.69} & \bestresult{62.19} & \secondresult{39.43} \\
 & \textbf{AsymLoop} & \bestresult{22.00} & \secondresult{50.99} & \secondresult{22.35} & \bestresult{54.50} & \bestresult{30.22} & \secondresult{38.38} & \secondresult{61.97} & \bestresult{40.06} \\
\cmidrule(lr){3-10}
\multirow{3}{*}{\textbf{350M}}
 & Dense & 22.20 & \bestresult{53.35} & \secondresult{26.62} & \secondresult{59.39} & 33.82 & \secondresult{39.15} & \secondresult{67.30} & \secondresult{43.12} \\
 & Full Loop & \bestresult{24.80} & 51.78 & 25.77 & 58.92 & \secondresult{34.57} & 39.05 & 66.38 & 43.04 \\
 & \textbf{AsymLoop} & \secondresult{23.60} & \secondresult{52.41} & \bestresult{27.05} & \bestresult{62.12} & \bestresult{35.59} & \bestresult{39.20} & \bestresult{67.79} & \bestresult{43.97} \\
\cmidrule(lr){3-10}
\multirow{3}{*}{\textbf{1B}}
 & Dense & 28.40 & \bestresult{59.19} & 33.96 & \secondresult{69.02} & 41.47 & 41.61 & \secondresult{71.49} & 49.31 \\
 & Full Loop & \secondresult{29.40} & 57.77 & \secondresult{34.64} & 68.64 & \secondresult{42.67} & \bestresult{42.53} & 71.38 & \secondresult{49.58} \\
& \textbf{AsymLoop}         & \bestresult{32.40} & \secondresult{57.93} & \bestresult{36.69} & \bestresult{70.71} & \bestresult{44.34} & \secondresult{41.81} & \bestresult{72.03} & \bestresult{50.85} \\
\midrule
\multicolumn{10}{c}{\textbf{Five-shot}} \\
\cmidrule(lr){1-10}
\multirow{3}{*}{\textbf{60M}}
 & Dense & \bestresult{17.60} & 49.57 & 18.43 & 42.93 & 27.09 & 34.54 & 58.11 & 35.47 \\
 & Full Loop & 15.40 & \secondresult{50.51} & \bestresult{20.48} & \secondresult{44.44} & \secondresult{27.27} & \secondresult{35.21} & \secondresult{58.32} & \secondresult{35.95} \\
 & \textbf{AsymLoop} & \secondresult{16.60} & \bestresult{51.70} & \secondresult{18.86} & \bestresult{45.71} & \bestresult{27.53} & \bestresult{36.13} & \bestresult{59.47} & \bestresult{36.57} \\
\cmidrule(lr){3-10}
\multirow{3}{*}{\textbf{135M}}
 & Dense & \secondresult{18.00} & \secondresult{51.70} & 22.35 & 50.21 & 28.75 & 36.44 & 61.64 & 38.44 \\
 & Full Loop & 17.20 & \bestresult{52.09} & \secondresult{22.44} & \secondresult{50.72} & \secondresult{28.85} & \bestresult{38.33} & \bestresult{63.06} & \secondresult{38.96} \\
 & \textbf{AsymLoop} & \bestresult{21.20} & 50.28 & \bestresult{23.46} & \bestresult{53.66} & \bestresult{30.03} & \secondresult{37.56} & \secondresult{62.95} & \bestresult{39.88} \\
\cmidrule(lr){3-10}
\multirow{3}{*}{\textbf{350M}}
 & Dense & \secondresult{23.00} & \secondresult{53.28} & \secondresult{27.65} & 59.93 & 33.41 & 39.00 & 66.76 & \secondresult{43.29} \\
 & Full Loop & 21.20 & 52.17 & 26.88 & \secondresult{60.02} & \secondresult{34.33} & \secondresult{39.15} & \secondresult{66.92} & 42.95 \\
 & \textbf{AsymLoop} & \bestresult{24.00} & \bestresult{54.38} & \bestresult{28.92} & \bestresult{64.48} & \bestresult{35.03} & \bestresult{40.69} & \bestresult{68.17} & \bestresult{45.10} \\
\cmidrule(lr){3-10}
\multirow{3}{*}{\textbf{1B}}
 & Dense & 24.20 & 57.70 & 36.35 & 69.57 & 41.07 & 42.48 & 71.60 & 48.99 \\
 & Full Loop & \secondresult{26.60} & \secondresult{58.25} & \secondresult{36.77} & \secondresult{70.54} & \secondresult{42.84} & \secondresult{43.09} & \bestresult{72.25} & \secondresult{50.05} \\
& \textbf{AsymLoop}         & \bestresult{29.00} & \bestresult{59.27} & \bestresult{38.99} & \bestresult{73.40} & \bestresult{44.06} & \bestresult{43.65} & \secondresult{71.76} & \bestresult{51.45} \\
\bottomrule
\end{tabular}
\end{table}

\subsection{Downstream Evaluation}
\label{subsec:downstream}

We evaluate the pretrained models on seven commonsense reasoning benchmarks to assess their downstream generalization: PIQA~\citep{bisk2020piqa}, Social IQa~\citep{sap2019socialiqa}, HellaSwag~\citep{zellers2019hellaswag}, WinoGrande~\citep{sakaguchi2021winogrande}, ARC-Challenge~\citep{clark2018think}, ARC-Easy~\citep{clark2018think}, and OpenBookQA~\citep{mihaylov2018can}. Evaluations cover all four model sizes under both zero-shot and five-shot settings, using the standard prompts provided by the LM Evaluation Harness~\citep{eval-harness}. Table~\ref{tab:downstream} summarizes the per-task and average accuracies.

As shown in Table~\ref{tab:downstream}, AsymLoop consistently outperforms Dense and Full Loop in average accuracy across all four model sizes under both zero-shot and five-shot settings. The improvements extend across multiple benchmarks, with particularly broad gains in five-shot evaluation: AsymLoop achieves the best performance on all seven tasks at 350M and six of the seven tasks at 1B. At 350M, its average accuracy exceeds Full Loop by 2.15 percentage points. These results demonstrate that the pre-training gains of AsymLoop translate into stronger downstream commonsense reasoning across both evaluation settings. We further analyze the training dynamics of downstream performance at the 1B scale in Appendix~\ref{app:downstream-trajectory-1b}.

Furthermore, we evaluate the mathematical reasoning performance of the 1B pretrained models through supervised fine-tuning on GSM8K~\citep{cobbe2021gsm8k}. AsymLoop achieves an accuracy of 11.75\%, a 1.52 percentage point improvement over Full Loop. These results suggest that its downstream benefits also extend to supervised fine-tuning; detailed results and training settings are provided in Appendix~\ref{app:gsm8k-sft}.

\subsubsection{Downstream Performance under a $5\times$ Chinchilla Budget}
\label{subsubsec:downstream-extended-budget}

To investigate whether the benefits of AsymLoop persist under extended training, we increase the training budget of the 350M models from 7B to 35B tokens, corresponding to five times the Chinchilla budget. We compare AsymLoop with Dense and Full Loop under the same token budget, maintaining matched effective depth between the recurrent models. Table~\ref{tab:downstream-35b} presents their zero-shot and five-shot performance on the seven commonsense reasoning benchmarks.

As shown in Table~\ref{tab:downstream-35b}, AsymLoop maintains the highest average accuracy in both zero-shot and five-shot evaluation under the extended training budget. In five-shot evaluation, it matches or surpasses both baselines on all seven tasks, improving average accuracy over Full Loop from 46.67\% to 47.41\%. These results demonstrate that the downstream benefits of AsymLoop persist with a fivefold increase in training tokens, supporting its effectiveness under extended pre-training.

\begin{table}[H]
\caption{\textbf{Zero- and five-shot downstream accuracy ($\uparrow$, $\%$) of 350M models trained on 35B tokens.} Avg. is the mean across seven tasks. Darker/lighter blue mark the best/second-best results within each shot setting.}
\label{tab:downstream-35b}
\centering
\papertablestyle
\setlength{\tabcolsep}{3pt}
\begin{tabular}{lrrrrrrrr}
\toprule
\textbf{Model} & \textbf{OBQA} & \textbf{WinoGrande} & \textbf{ARC-c} & \textbf{ARC-e} & \textbf{HellaSwag} & \textbf{SIQA} & \textbf{PIQA} & \textbf{Avg.} \\
\midrule
\multicolumn{9}{c}{\textbf{Zero-shot}} \\
\cmidrule(lr){1-9}
Dense & 25.40 & \secondresult{54.70} & 31.06 & \secondresult{66.33} & 37.69 & 39.61 & 68.44 & 46.18 \\
Full Loop ($K=2$) & \bestresult{27.20} & 53.43 & \secondresult{31.91} & 66.20 & \bestresult{39.04} & \bestresult{41.40} & \secondresult{68.55} & \secondresult{46.82} \\
\textbf{AsymLoop} ($K=3$, ALRS) & \secondresult{26.80} & \bestresult{55.80} & \bestresult{32.68} & \bestresult{68.35} & \secondresult{38.54} & \secondresult{40.02} & \bestresult{69.04} & \bestresult{47.32} \\
\midrule
\multicolumn{9}{c}{\textbf{Five-shot}} \\
\cmidrule(lr){1-9}
Dense & 21.20 & 54.54 & 30.72 & \secondresult{67.00} & 37.21 & \secondresult{41.20} & \secondresult{68.77} & 45.81 \\
Full Loop ($K=2$) & \secondresult{23.00} & \secondresult{56.51} & \secondresult{31.66} & 66.50 & \secondresult{38.53} & \bestresult{41.50} & \bestresult{68.99} & \secondresult{46.67} \\
\textbf{AsymLoop} ($K=3$, ALRS) & \bestresult{25.40} & \bestresult{56.67} & \bestresult{32.08} & \bestresult{68.56} & \bestresult{38.70} & \bestresult{41.50} & \bestresult{68.99} & \bestresult{47.41} \\
\bottomrule
\end{tabular}
\end{table}

\subsection{Effect of Recurrent Depth}
\label{subsec:recurrence-validation}

\begin{table*}[!htb]
\caption{\textbf{Effect of recurrent depth.} Validation perplexity ($\downarrow$) of Dense, Full Loop, and AsymLoop across model sizes at matched effective depths. Darker/lighter blue mark lower/higher perplexity within each depth-matched pair.}
\label{tab:recurrence-validation}
\centering
\papertablestyle
\begin{tabular}{lccclcc}
\toprule
\textbf{Scale} & \shortstack{\textbf{Loop / Total}\\\textbf{Layers}} & \textbf{$K$} & \shortstack{\textbf{Effective}\\\textbf{Depth}} & \textbf{Architecture} & \textbf{Budget} & \textbf{PPL} \\
\midrule
\multirow{6}{*}{\textbf{60M}}
 & 0 / 6 & 1 & 6  & Dense                    & 1.68B & \secondresult{23.72} \\
 & 3 / 6 & 1 & 6  & \textbf{AsymLoop (Ours)} & 1.68B & \bestresult{21.79} \\
\cmidrule(lr){2-7}
 & 6 / 6 & 2 & 12 & Full Loop                & 1.68B & \secondresult{23.03} \\
 & 3 / 6 & 3 & 12 & \textbf{AsymLoop (Ours)} & 1.68B & \bestresult{21.45} \\
\cmidrule(lr){2-7}
 & 6 / 6 & 3 & 18 & Full Loop                & 1.68B & \secondresult{22.77} \\
 & 3 / 6 & 5 & 18 & \textbf{AsymLoop (Ours)} & 1.68B & \bestresult{21.30} \\
\midrule
\multirow{6}{*}{\textbf{135M}}
 & 0 / 12 & 1 & 12 & Dense                    & 3.00B & \secondresult{17.46} \\
 & 6 / 12 & 1 & 12 & \textbf{AsymLoop (Ours)} & 3.00B & \bestresult{16.54} \\
\cmidrule(lr){2-7}
 & 12 / 12 & 2 & 24 & Full Loop                & 3.00B & \secondresult{17.12} \\
 & 6 / 12  & 3 & 24 & \textbf{AsymLoop (Ours)} & 3.00B & \bestresult{16.14} \\
\cmidrule(lr){2-7}
 & 12 / 12 & 3 & 36 & Full Loop                & 3.00B & \secondresult{16.81} \\
 & 6 / 12  & 5 & 36 & \textbf{AsymLoop (Ours)} & 3.00B & \bestresult{16.09} \\
\midrule
\multirow{6}{*}{\textbf{350M}}
 & 0 / 24  & 1 & 24 & Dense                    & 7.00B & \secondresult{12.55} \\
 & 12 / 24 & 1 & 24 & \textbf{AsymLoop (Ours)} & 7.00B & \bestresult{12.20} \\
\cmidrule(lr){2-7}
 & 24 / 24 & 2 & 48 & Full Loop                & 7.00B & \secondresult{12.30} \\
 & 12 / 24 & 3 & 48 & \textbf{AsymLoop (Ours)} & 7.00B & \bestresult{12.00} \\
\cmidrule(lr){2-7}
 & 24 / 24 & 3 & 72 & Full Loop                & 7.00B & \secondresult{12.15} \\
 & 12 / 24 & 5 & 72 & \textbf{AsymLoop (Ours)} & 7.00B & \bestresult{11.92} \\
\midrule
\multirow{2}{*}{\textbf{1B}}
 & 24 / 24 & 2 & 48 & Full Loop                & 20.00B & \secondresult{8.91} \\
 & 12 / 24 & 3 & 48 & \textbf{AsymLoop (Ours)} & 20.00B & \bestresult{8.86} \\
\bottomrule
\end{tabular}
\end{table*}

We investigate how recurrent depth affects AsymLoop by varying the number of decoder iterations from $K=1$ to $K=3$ and $K=5$ at the 60M, 135M, and 350M scales. To compare architectures at matched effective depth, these configurations are paired with Dense, Full Loop with $K=2$, and Full Loop with $K=3$, respectively, yielding depths of $L$, $2L$, and $3L$. The training token budget is held fixed within each model scale. Table~\ref{tab:recurrence-validation} reports validation perplexity for these comparisons.

As shown in Table~\ref{tab:recurrence-validation}, AsymLoop achieves lower validation perplexity than the corresponding depth-matched baseline in every comparison. Its advantage is already evident at $K=1$ and persists as effective depth increases to $2L$ and $3L$. Increasing the number of decoder iterations further reduces perplexity across the 60M, 135M, and 350M scales. For example, at 350M, perplexity decreases from 12.20 at $K=1$ to 11.92 at $K=5$. These results demonstrate that AsymLoop remains effective across different depths and benefits from additional recurrent computation.

\subsection{Ablation Studies and Analysis}
\label{subsec:ablations}

We examine the contributions of AsymLoop's architectural and optimization components. Further analyses of asymmetric learning-rate allocation, Global KV under recurrence, and alternative context-access designs are provided in Appendices~\ref{app:alrs-diagnostics} and~\ref{app:ablation-details}.

\subsubsection{Component Ablations}

We evaluate the contributions of AsymLoop's attention design, normalization, and learning-rate allocation through component ablations (Table~\ref{tab:ablations}). For the architectural components, we individually remove Global KV, Local KV, and QKNorm under uniform learning rates, and include a parameter-matched variant without Global KV. For optimization, we compare ALRS with uniform learning rates and a fixed decoder learning-rate boost, and examine sensitivity to the base learning rate.

\begin{table}[!htbp]
\caption{\textbf{Ablations at the 60M scale.} Validation perplexity ($\downarrow$). \textbf{(a)} Architectural components and \textbf{(b)} learning-rate ablations on AsymLoop ($K=3,5$). \textbf{(c)} Uniform LR versus ALRS on Full Loop ($K=2$). Darker/lighter blue mark the best/second-best results within each panel of (a, b).}
\label{tab:ablations}
\centering
\papertablestyle
\setlength{\tabcolsep}{10pt}
\begin{tabular}{lrr}
\toprule
\textbf{Model Variant} & \textbf{PPL $\downarrow$ ($K=3$)} & \textbf{PPL $\downarrow$ ($K=5$)} \\
\midrule
\multicolumn{3}{l}{\textit{(a) Architectural components}} \\[2pt]
\textbf{AsymLoop (ALRS)} & \bestresult{23.751} & \bestresult{23.580} \\
\addlinespace[2pt]
\textbf{AsymLoop (uniform LR)} & \secondresult{24.727} & \secondresult{24.619} \\
\quad \textit{w/o} QKNorm & 24.984 & 24.663 \\
\quad \textit{w/o} Global KV & 25.566 & 25.751 \\
\quad \textit{w/o} Local KV & 25.656 & 25.465 \\
\quad Baseline with parameter matching & 25.339 & 25.323 \\
\midrule
\multicolumn{3}{l}{\textit{(b) ALRS and base learning rate}} \\[2pt]
\textbf{AsymLoop} & \bestresult{23.751} & \bestresult{23.580} \\
\quad \textit{w/o} ALRS & 24.727 & 24.619 \\
\quad \textit{w/} $0.5\times$ LR & \secondresult{24.059} & \secondresult{23.826} \\
\quad \textit{w/} $2\times$ LR & 24.865 & 24.278 \\
\quad \textit{w/} $4\times$ LR & 26.181 & 25.941 \\
\quad \textit{w/} static decoder boost & 24.785 & 24.708 \\
\midrule
\textit{(c) Full Loop ($K=2$)} & \textbf{Uniform LR} & \textbf{ALRS} \\[2pt]
Validation PPL $\downarrow$ & 25.571 & 24.787 \\
\bottomrule
\end{tabular}
\end{table}

\textbf{Architectural components.} Under uniform learning rates, removing either Global KV or Local KV increases validation perplexity at both $K=3$ and $K=5$ (Table~\ref{tab:ablations}a), indicating that encoder context and evolving decoder states provide complementary information. Retaining Global KV also yields lower perplexity than reallocating its parameters to a wider FFN, suggesting that access to encoder context is more effective than additional FFN capacity in these comparisons. Removing QKNorm increases perplexity at both recurrent depths.

\textbf{ALRS and base learning rate.} At the default base learning rate, ALRS lowers perplexity by approximately one point relative to uniform learning rates at both recurrent depths (Table~\ref{tab:ablations}b). It also achieves lower perplexity than a fixed decoder learning-rate boost, indicating the benefit of adapting learning rates at the matrix level. Among the tested base learning rates, $0.001$ gives the best results at both $K=3$ and $K=5$.

AsymLoop also retains its advantage over a widened Full Loop baseline at matched analytical per-step training FLOPs; detailed comparisons are provided in Appendix~\ref{app:flops-matched-results}.

\section{Conclusion}
\label{sec:conclusion}

We presented AsymLoop, an asymmetric looped Transformer that combines single-pass context encoding with recurrent state refinement through joint attention to Global and Local KV. We train this architecture using the spectral layerwise learning-rate method of \citet{he2026one} and evaluate its effect through optimization ablations. Across four model scales from 60M to 1B, AsymLoop achieves lower validation perplexity and higher average downstream accuracy than the depth-matched Full Loop baseline. Its language-modeling gains extend to an OLMo-style backbone and sparse MoE feed-forward layers, while supervised fine-tuning on GSM8K also yields higher mathematical reasoning accuracy. Although AsymLoop requires more training FLOPs than the original Full Loop, it achieves lower validation loss than a widened Full Loop baseline under matched training FLOPs.
\newpage
\subsection*{AI use statement}

Generative AI tools assisted with polishing the language of this manuscript. The authors reviewed the AI-assisted text and take responsibility for all content and claims in this paper.

\subsection*{Reproducibility statement}

The model architecture is described in Section~\ref{sec:method}, and Appendix~\ref{app:reproducibility-details} provides the training configurations, hyperparameters, and computational resources. All source datasets used in our experiments are publicly available; Section~\ref{subsec:exp_setup} and the relevant appendices specify our FineWeb-Edu shard selection and other custom training and validation splits. For the seven downstream benchmarks, we use the standard prompts and evaluation protocols of the LM Evaluation Harness~\citep{eval-harness}; the separate GSM8K protocol is documented in Appendix~\ref{app:gsm8k-sft}. The H100 pre-training runs used up to 32 NVIDIA H100 GPUs per run, while hardware for other experiments is identified in their respective protocols. We plan to publicly release the training and evaluation code and the model checkpoints used in this work.

\bibliography{references}
\bibliographystyle{preprint}

\clearpage
\appendix
\setlength{\intextsep}{8pt plus 2pt minus 2pt}
\section{Architecture, Training, and Evaluation Details}
\label{app:reproducibility-details}

Table~\ref{tab:hyperparams} summarizes the architectures and pre-training hyperparameters for the four model configurations. We denote these configurations by 60M, 135M, 350M, and 1B, with the actual parameter counts reported separately for each architecture.

All main pre-training experiments use cleaned and deduplicated FineWeb-Edu shards 00000--00039 for training, with shard 00233 reserved for validation. We use the t5-base tokenizer with a vocabulary size of 32,128 and a sequence length of 1024. Input embeddings and the output head are untied.

Training budgets are matched across architectures within each model configuration. The 60M models are trained for 3,200 optimizer steps, corresponding to approximately 1.68B input token positions including padding. The 135M, 350M, and 1B models are trained on 3B, 7B, and 20B tokens, respectively.

\begin{table}[H]
\caption{\textbf{Model architectures and pre-training hyperparameters.} Backbone configurations follow~\citet{he2026one}. Parameter counts include the untied input embeddings and output head.}
\label{tab:hyperparams}
\centering
\papertablestyle
\begin{tabular}{lcccc}
\toprule
\textbf{Configuration} & \textbf{60M} & \textbf{135M} & \textbf{350M} & \textbf{1B} \\
\midrule
Dense parameters & 91.8M & 134.3M & 368.2M & 1.34B \\
AsymLoop parameters & 95.4M & 141.4M & 393.4M & 1.44B \\
Layers $L$ ($E+D$) & 6 (3+3) & 12 (6+6) & 24 (12+12) & 24 (12+12) \\
$d_{\text{model}}$ & 768 & 768 & 1024 & 2048 \\
$d_{\text{ff}}$ & 2048 & 2048 & 2736 & 5461 \\
Heads & 6 & 12 & 16 & 32 \\
Training budget & 1.68B & 3.00B & 7.00B & 20.00B \\
Peak LR $\eta_0$ & $1.0\times10^{-3}$ & $1.0\times10^{-3}$ & $1.0\times10^{-3}$ & $5.0\times10^{-4}$ \\
Weight decay & 0.1 & 0.1 & 0.1 & 0.1 \\
Effective depth ($L/2L/3L$) & 6 / 12 / 18 & 12 / 24 / 36 & 24 / 48 / 72 & 24 / 48 / 72 \\
\bottomrule
\end{tabular}
\end{table}

We use AdamW with $(\beta_1,\beta_2)=(0.9,0.95)$, $\epsilon=10^{-8}$, and weight decay 0.1. All main pre-training experiments use a global batch size of 512, a global gradient clipping threshold of 1.0 in L2 norm, and random seed 6198. The base learning rate follows linear warmup over the first 10\% of training, followed by cosine decay to 10\% of its peak. Schedule progress is measured in optimizer steps at 60M and training tokens at larger scales. ALRS scheduling details are provided in Appendix~\ref{app:alrs-schedule}.

Large-scale pre-training uses up to 32 NVIDIA H100 GPUs, each with 80 GB of memory.

We report validation cross-entropy (CE) and perplexity (PPL) to assess pre-training performance. For downstream commonsense reasoning, we report zero-shot and five-shot accuracy using the standard prompts provided by the LM Evaluation Harness~\citep{eval-harness}.

\section{ALRS Allocation and Weight-Spectrum Diagnostics}
\label{app:alrs-diagnostics}

\subsection{ALRS Scheduling Details}
\label{app:alrs-schedule}

Following the allocation and scheduling procedure of \citet{he2026one}, ALRS assigns matrix-wise learning-rate multipliers in $[1,5]$. Allocations are updated at the first optimizer step and every 100 steps thereafter during the first 20\% of training, with a 50-step linear transition to each new learning-rate target. After this initial phase, the multipliers remain fixed while the base learning rate continues its cosine decay.

\subsection{Observed Encoder--Decoder Learning-Rate Differences}
\label{subsec:alrs-allocation-analysis}

\begin{figure}[H]
    \centering
    \includegraphics[width=0.98\linewidth]{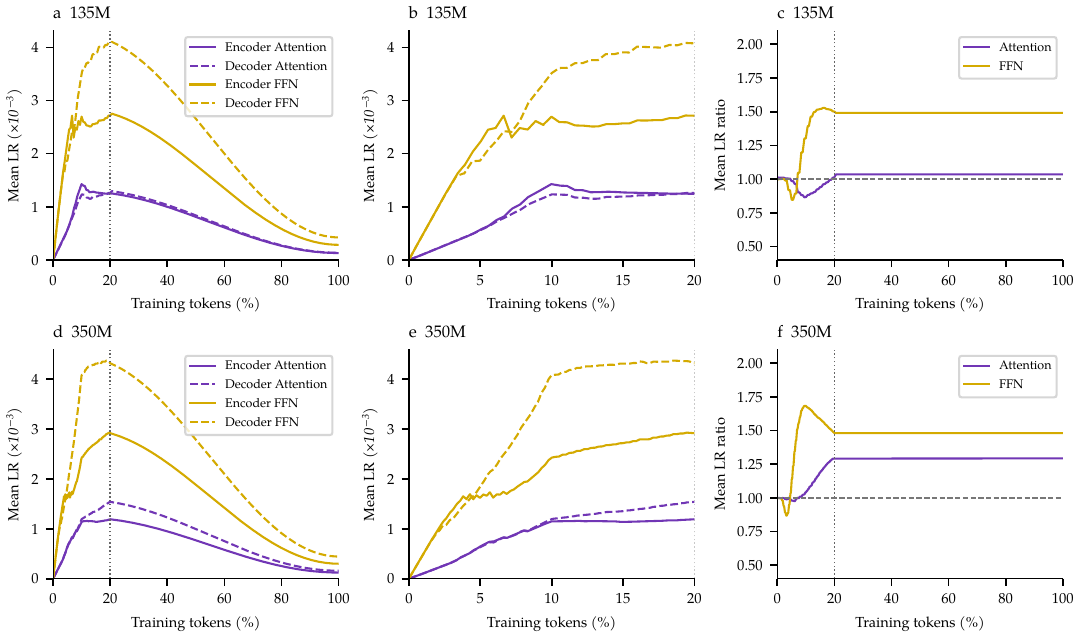}
    \caption{\textbf{Learning-rate dynamics under ALRS at 135M and 350M.} \textbf{Left:} Full training trajectories. \textbf{Middle:} The first 20\% of training. \textbf{Right:} Decoder-to-encoder mean learning-rate ratios. Vertical dotted lines mark the end of allocation updates.}
    \label{fig:alrs-trajectory}
\end{figure}

We analyze how ALRS allocates learning rates between the encoder and decoder in the 135M and 350M models. Figure~\ref{fig:alrs-trajectory} shows their mean learning-rate trajectories and decoder-to-encoder ratios separately for Attention and FFN modules.

Figure~\ref{fig:alrs-trajectory} shows a consistent FFN asymmetry across model scales: after the allocation updates end, the decoder's mean FFN learning rate is approximately 1.5 times the encoder's. For Attention, the encoder--decoder difference is more pronounced at 350M than at 135M. These differences arise from applying the same spectral allocation rule to all matrices; the approximately 1.5-fold FFN ratio is an observed outcome rather than a prescribed decoder multiplier.

\subsection{Weight-Spectrum Analysis}
\label{app:alrs-spectrum-details}

To examine the weight structure learned under ALRS, we compare normalized stable rank between uniform learning rates and ALRS in the 135M and 350M AsymLoop models ($K=3$). Statistics are averaged equally over the local Q/K/V/O matrices for Attention and the gate/up/down matrices for FFN, separately for the encoder and decoder.

\begin{figure}[H]
    \centering
    \includegraphics[width=0.92\textwidth]{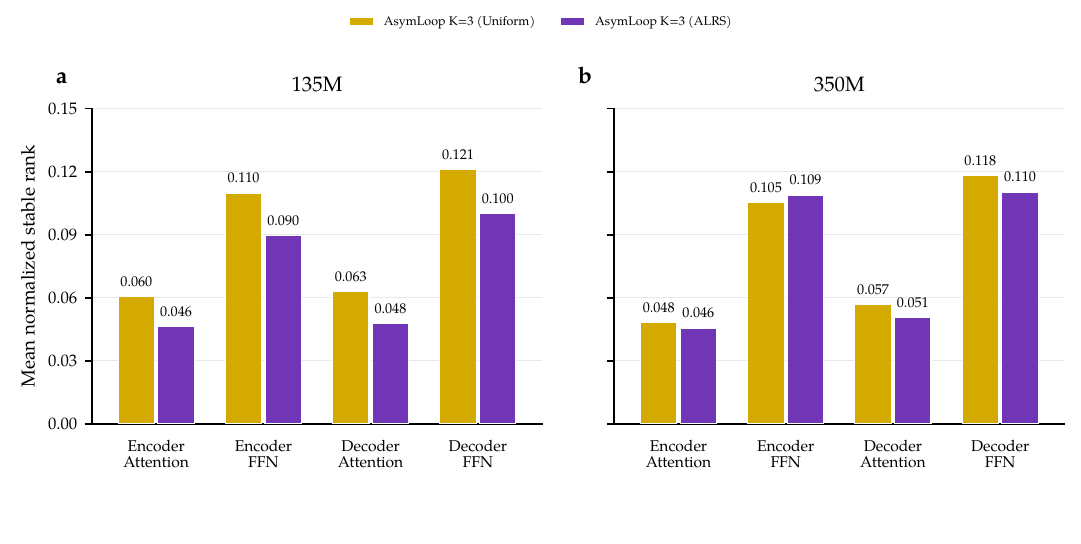}
    \caption{\textbf{Normalized stable rank under uniform learning rates and ALRS.} Results for 135M and 350M AsymLoop models ($K=3$). For each matrix $W$, we compute $\|W\|_F^2/[\|W\|_2^2\min(\mathrm{shape}(W))]$ and average equally within each module group.}
    \label{fig:alrs-stable-rank}
\end{figure}

Figure~\ref{fig:alrs-stable-rank} shows that ALRS generally produces lower normalized stable rank, with consistent reductions in Attention and decoder FFN modules at both model scales. This pattern reflects greater relative concentration of spectral energy in the leading singular direction, indicating that differentiated learning-rate allocation is accompanied by changes in the learned weight structure.

\section{Complete Ablation Results and Evaluation Details}
\label{app:ablation-details}

\subsection{Global KV under Recurrence}
\label{subsec:kv-interaction}

We compare AsymLoop with and without Global KV at $K=1,3,5$ to examine how its contribution changes with recurrent depth. Both variants retain QKNorm and use uniform learning rates.

\begin{table}[H]
\centering
\papertablestyle
\setlength{\tabcolsep}{10pt}
\caption{\textbf{Global KV and recurrent depth.} Validation CE ($\downarrow$) with and without Global KV. $G(K)$ is the CE reduction from Global KV; the final row reports the additional gain $G(K)-G(1)$.}
\label{tab:kv-interaction}
\begin{tabular}{lrrr}
\toprule
\textbf{Model Variant} & \textbf{CE $\downarrow$ ($K=1$)} & \textbf{CE $\downarrow$ ($K=3$)} & \textbf{CE $\downarrow$ ($K=5$)} \\
\midrule
\textbf{AsymLoop} & \bestresult{3.238735} & \bestresult{3.207915} & \bestresult{3.203515} \\
\quad \textit{w/o} Global KV & \secondresult{3.260346} & \secondresult{3.241257} & \secondresult{3.248483} \\
\midrule
Global KV gain $G(K)$ & 0.021611 & 0.033343 & 0.044968 \\
Gain beyond $K=1$ & 0.000000 & 0.011732 & 0.023357 \\
\bottomrule
\end{tabular}
\end{table}

Global KV reduces validation cross-entropy at every tested depth, with its gain increasing from 0.022 at $K=1$ to 0.045 at $K=5$ (Table~\ref{tab:kv-interaction}). This trend supports a greater benefit from encoder context as recurrence deepens.

\subsection{Alternative Context-Access Designs}
\label{subsec:strengthened-controls}

We compare Global KV with two alternatives: removing this access path or replacing it with additive injection of the encoder output before each decoder iteration. The additive variant computes $z_0=0$ and $z_t=\mathcal{D}(z_{t-1}+E)$, where $E$ is the encoder output. Full Loop provides an additional baseline. All models use QKNorm and uniform learning rates.

\begin{table}[H]
\centering
\papertablestyle
\setlength{\tabcolsep}{10pt}
\caption{\textbf{Context-access variants and a Full Loop baseline.} Validation perplexity ($\downarrow$). Full Loop uses $K=2$; all other variants use $K=3$. Darker/lighter blue mark the best/second-best results.}
\label{tab:strengthened-controls}
\begin{tabular}{lr}
\toprule
\textbf{Model Variant} & \textbf{PPL $\downarrow$} \\
\midrule
\textbf{AsymLoop} & \bestresult{24.727} \\
\quad \textit{w/o} Global KV & 25.566 \\
\quad \textit{w/} additive injection & 25.116 \\
\midrule
Full Loop & \secondresult{25.053} \\
\bottomrule
\end{tabular}
\end{table}

As shown in Table~\ref{tab:strengthened-controls}, AsymLoop achieves the lowest validation perplexity. Additive injection improves over removing Global KV, while retaining Global KV provides a further gain. These results highlight the benefit of accessing encoder context through attention during recurrent computation.

\subsection{Additional Validation Results}

Table~\ref{tab:strict-references} reports validation cross-entropy and perplexity for Dense, Full Loop, and AsymLoop without QKNorm under a common evaluation protocol. The AsymLoop results also provide the reference values for the QKNorm ablation in Table~\ref{tab:ablations}.

\begin{table}[H]
\centering
\papertablestyle
\caption{\textbf{Additional validation results.} Cross-entropy and perplexity ($\downarrow$) under a common evaluation protocol. Darker/lighter blue mark the best/second-best results in each metric column.}
\label{tab:strict-references}
\begin{tabular}{lrrrr}
\toprule
Reference & $K$ & Effective depth & CE & PPL \\
\midrule
Dense, Uniform & 1 & 6 & 3.271820 & 26.359274 \\
Dense, layerwise LR & 1 & 6 & \secondresult{3.209192} & \secondresult{24.759083} \\
Full Loop, Uniform & 2 & 12 & 3.241455 & 25.570904 \\
Full Loop, Uniform & 3 & 18 & 3.229346 & 25.263120 \\
AsymLoop w/o QKNorm, Uniform & 3 & 12 & 3.218239 & 24.984094 \\
AsymLoop w/o QKNorm, Uniform & 5 & 18 & \bestresult{3.205305} & \bestresult{24.663015} \\
\bottomrule
\end{tabular}
\end{table}

\section{Computational Cost and Downstream Performance}
\label{app:computational-cost}

We analyze AsymLoop's computational cost and its relationship with downstream performance. Compute-matched comparisons with a widened Full Loop baseline are reported in Appendix~\ref{app:flops-matched-results}.

\subsection{Cost of Global KV and Joint Attention}
\label{app:cost-decomposition}

Let $N,d,m,$ and $V$ denote the sequence length, hidden dimension, SwiGLU intermediate dimension, and vocabulary size, respectively. We count two FLOPs per multiply--accumulate and use full attention matrices. A standard Transformer block has forward cost
\begin{equation}
    C(m)=8Nd^2+6Ndm+4N^2d.
    \label{eq:cost-standard-block}
\end{equation}

At matched effective depth $J=E+KD=RL$, the Full Loop cost is $F_{\mathrm{Full}}=F_{\mathrm{blk}}+F_{\mathrm{head}}$, where $F_{\mathrm{blk}}=JC(m)$ and $F_{\mathrm{head}}=2NdV$. Here, $E=D=L/2$ and $R=(K+1)/2$.

AsymLoop adds Global KV projections computed once per decoder layer and additional attention products at each decoder iteration:
\begin{equation}
    \begin{aligned}
        F_{\mathrm{proj}}&=4DNd^2,\qquad F_{\mathrm{glob}}=4KDN^2d,\\
        F_{\mathrm{Asym}}&=F_{\mathrm{Full}}+F_{\mathrm{proj}}+F_{\mathrm{glob}}.
    \end{aligned}
    \label{eq:cost-asym-overhead}
\end{equation}

\begin{table}[H]
\centering
\papertablestyle
\caption{\textbf{Analytical forward FLOPs at matched effective depth.} Full Loop uses $R=2$ and AsymLoop uses $K=3$, with $N=1024$ and $V=32128$. Costs are reported per input position; overhead is relative to Full Loop.}
\label{tab:analytical-forward-cost}
\begin{tabular}{lrrrrr}
\toprule
& \multicolumn{2}{c}{Forward GFLOPs/token} & \multicolumn{3}{c}{Additional cost (\%)} \\
\cmidrule(lr){2-3}\cmidrule(lr){4-6}
Model size & Full Loop & AsymLoop & Global KV proj. & Global attn. & Total \\
\midrule
60M  & 0.256967 & 0.292356 & 2.75 & 11.02 & 13.77 \\
135M & 0.464585 & 0.535364 & 3.05 & 12.19 & 15.23 \\
350M & 1.476657 & 1.677984 & 3.41 & 10.23 & 13.63 \\
1B   & 5.365891 & 5.869208 & 3.75 & 5.63 & 9.38 \\
\bottomrule
\end{tabular}
\end{table}

Table~\ref{tab:analytical-forward-cost} summarizes the resulting forward costs. Across the evaluated model sizes, these components add 9.38\%--15.23\% to the Full Loop baseline.

\subsection{Compute--Performance Trade-offs Across Recurrent Depths}
\label{app:observed-compute-performance}

\begin{figure}[H]
\centering
\includegraphics[width=\linewidth]{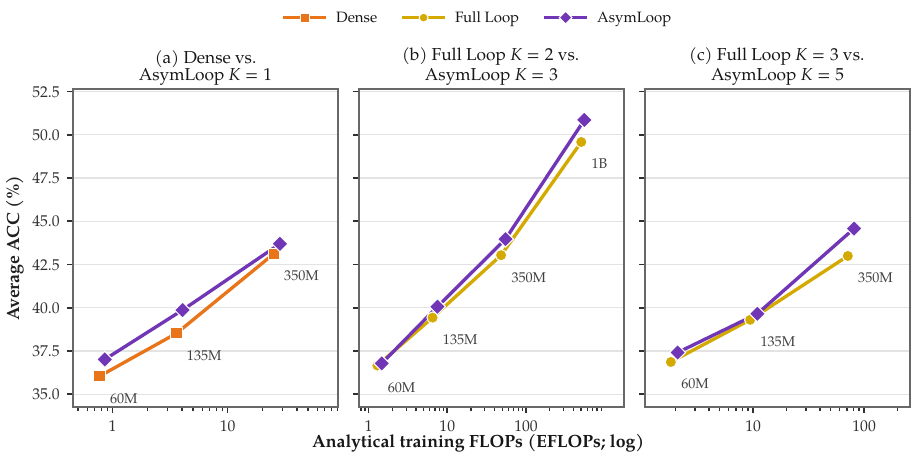}
\caption{\textbf{Training compute--performance trade-offs across model scales and recurrent depths.} Average zero-shot accuracy over seven downstream tasks versus analytical training FLOPs. The horizontal axis uses a logarithmic scale in EFLOPs ($10^{18}$ FLOPs). Panels pair AsymLoop with the baseline of matched effective depth: (a) Dense and AsymLoop $K=1$ ($L$), (b) Full Loop $K=2$ and AsymLoop $K=3$ ($2L$), and (c) Full Loop $K=3$ and AsymLoop $K=5$ ($3L$). Purple, yellow, and orange denote AsymLoop, Full Loop, and Dense, respectively.}
\label{fig:training-compute-pareto}
\end{figure}

We investigate the trade-off between training compute and downstream performance across model scales and recurrent depths. Figure~\ref{fig:training-compute-pareto} compares average zero-shot accuracy against analytical training FLOPs, pairing AsymLoop with Dense or Full Loop at matched effective depths of $L$, $2L$, and $3L$.

Across all three depth settings, AsymLoop consistently achieves higher accuracy than its depth-matched baseline with a modest increase in training compute. Comparing across depths further reveals a favorable compute--performance trade-off: at 60M, 135M, and 350M, AsymLoop with $K=1$ achieves higher accuracy than Full Loop with either $K=2$ or $K=3$, while requiring fewer training FLOPs. These results show that AsymLoop can offer both better downstream performance and lower compute than deeper Full Loop configurations. Complementary comparisons with a widened Full Loop baseline at matched analytical per-step training FLOPs are provided in Appendix~\ref{app:flops-matched-results}.

\section{Performance at Matched Per-Step Training FLOPs}
\label{app:flops-matched-results}

To investigate whether AsymLoop's performance gains arise solely from additional training FLOPs, we compare it with a widened Full Loop baseline. We increase the baseline's SwiGLU intermediate dimension while preserving its recurrent architecture, matching AsymLoop's analytical per-step training FLOPs under a full-block recomputation assumption.

\subsection{Experimental Setup}
\label{app:flops-matched-protocol}

We evaluate the original Full Loop, widened Full Loop, and AsymLoop at the 60M and 135M scales. At both scales, the widened baseline increases the FFN dimension from 2048 to 2656. Full Loop uses two iterations, while AsymLoop uses three decoder iterations with ALRS. Within each scale, all models share the training data, token budget, batch size, and distributed training configuration. At 60M, we report validation cross-entropy from one run per configuration; at 135M, we report the mean and sample standard deviation over three training seeds per configuration.

\subsection{Results}
\label{app:flops-matched-main}

Table~\ref{tab:flops-matched} shows that widening Full Loop improves validation cross-entropy, closing approximately 40\% of its gap to AsymLoop at both model scales. Nevertheless, AsymLoop achieves the lowest cross-entropy in both comparisons. Its advantage also persists across all three random seeds in the 135M experiments. These results suggest that additional training FLOPs alone do not explain AsymLoop's performance gains.

At 135M, each configuration is evaluated over three training seeds. The widened Full Loop and AsymLoop achieve validation cross-entropies of $2.78983 \pm 0.00197$ and $2.74947 \pm 0.00299$, respectively (mean $\pm$ sample standard deviation). The mean difference is $0.04036$, and all three AsymLoop runs achieve lower cross-entropy than all three widened Full Loop runs. These observations support the consistency of the improvement within the evaluated training configuration.

\begin{table}[H]
\centering
\papertablestyle
\caption{\textbf{Validation cross-entropy and training cost with a widened Full Loop baseline.} The widened baseline matches AsymLoop's analytical per-step training FLOPs. The 60M results are single runs; the 135M results are means $\pm$ sample standard deviations over three training seeds. Step time and peak memory are measured on eight H100 GPUs. Darker/lighter blue marks the best/second-best cross-entropy at each model scale.}
\label{tab:flops-matched}
\begin{tabular}{llrrrrr}
\toprule
Model size & Model & FFN & Params & s/step & Peak mem. & Validation CE \\
\midrule
\multirow{3}{*}{60M}
 & Full Loop $K{=}2$        & 2048 & 91.83M  & 0.232 & 20.2\,GiB & 3.0434 \\
 & \quad + FLOPs matching   & 2656 & 100.23M & 0.230 & 21.6\,GiB & \secondresult{3.0139} \\
 & AsymLoop $K{=}3$         & 2048 & 95.37M  & 0.216 & 21.7\,GiB & \bestresult{2.9684} \\
\midrule
\multirow{3}{*}{135M}
 & Full Loop $K{=}2$        & 2048 & 134.30M & 0.286 & 30.9\,GiB & $2.81550 \pm 0.00221$ \\
 & \quad + FLOPs matching   & 2656 & 151.11M & 0.296 & 33.6\,GiB & \secondresult{$2.78983 \pm 0.00197$} \\
 & AsymLoop $K{=}3$         & 2048 & 141.38M & 0.367 & 33.8\,GiB & \bestresult{$2.74947 \pm 0.00299$} \\
\bottomrule
\end{tabular}
\end{table}

\subsection{Training FLOPs Verification}
\label{app:flops-matched-measured}

We measure per-step training matrix FLOPs on H100 GPUs, including forward, backward, and activation recomputation. Table~\ref{tab:flops-matched-measured} shows that the widened Full Loop uses 1.23\%--2.29\% fewer FLOPs than AsymLoop, indicating closely matched training compute.

\begin{table}[H]
\centering
\papertablestyle
\caption{\textbf{Measured matrix FLOPs per training step.} Counts include forward, backward, and activation recomputation, with sequence length 1024 and batch size 1.}
\label{tab:flops-matched-measured}
\begin{tabular}{lrrrr}
\toprule
& \multicolumn{2}{c}{Measured GFLOPs/step} & \multicolumn{2}{c}{Widened Full Loop} \\
\cmidrule(lr){2-3}\cmidrule(lr){4-5}
Model size & AsymLoop & Full Loop (wide) & FFN Width & Difference vs. AsymLoop \\
\midrule
60M  & 1134.878 & 1108.907 & 2656 & $-2.29\%$ \\
135M & 2118.157 & 2092.198 & 2656 & $-1.23\%$ \\
\bottomrule
\end{tabular}
\end{table}

\subsection{Discussion}
\label{app:flops-matched-boundaries}

AsymLoop achieves lower validation cross-entropy than the widened Full Loop at both model scales while using fewer parameters. Table~\ref{tab:flops-matched} also shows comparable peak memory usage and per-step training times of the same order of magnitude. These results highlight AsymLoop's performance advantage at closely matched training FLOPs.

\section{Downstream Training Dynamics at 1B}
\label{app:downstream-trajectory-1b}

\begin{figure}[H]
\centering
\includegraphics[width=\linewidth]{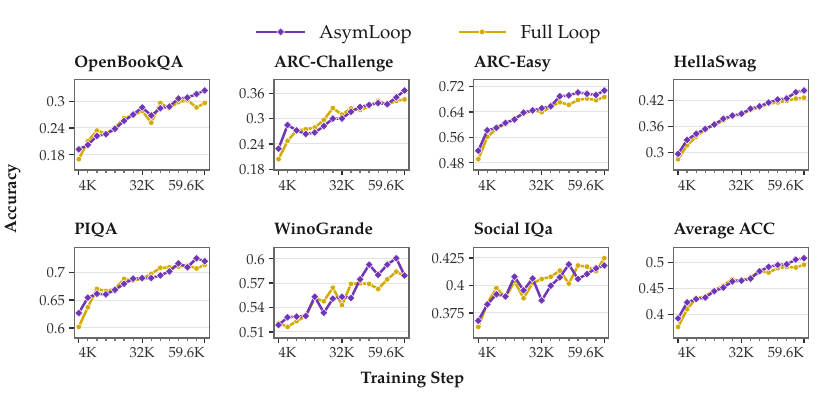}
\caption{\textbf{Downstream performance during pre-training at 1B.} Zero-shot accuracy on seven benchmarks and their average. Purple and yellow denote AsymLoop and Full Loop, respectively.}
\label{fig:downstream_trajectory_1b}
\end{figure}

We track the zero-shot performance of 1B Full Loop and AsymLoop throughout pre-training. Figure~\ref{fig:downstream_trajectory_1b} presents the accuracy trajectories on seven downstream benchmarks and their average.

Both models improve with training, while AsymLoop develops a clearer advantage in average accuracy during the later stages. This trend is particularly evident on OpenBookQA, ARC-Easy, and HellaSwag, where AsymLoop maintains higher accuracy across successive evaluations toward the end of training.

\section{Transfer to an OLMo-Style Backbone}
\label{app:olmo-transfer}

To assess generalization across backbone architectures, we evaluate AsymLoop on an OLMo-style backbone~\citep{groeneveld-etal-2024-olmo}. We compare Dense, Full Loop, and AsymLoop trained from scratch.

\subsection{Experimental Setup}
\label{app:olmo-protocol}

\paragraph{Architecture.}
All models use eight physical Transformer blocks with hidden dimension 512, eight attention heads, and SwiGLU intermediate dimension 2048, together with RMSNorm and RoPE. Full Loop repeats the backbone twice, while AsymLoop combines a four-layer encoder with a four-layer decoder repeated three times. The two looped models therefore have the same effective depth of 16.

\paragraph{Data and evaluation.}
We train all models on approximately 1.68B tokens from FineWeb-Edu-dedup using the SmolLM tokenizer, with sequence length 1024 and global batch size 512. Validation cross-entropy and perplexity are evaluated on a shared held-out set.

\paragraph{Optimization.}
We use AdamW with a cosine learning-rate schedule and 10\% warmup. AsymLoop applies ALRS multipliers in $[1,3]$, with allocation updates restricted to the first 20\% of training.

\subsection{Validation Results}
\label{app:olmo-results}

\begin{table}[H]
\caption{\textbf{Validation performance on an OLMo-style backbone.} Cross-entropy and perplexity ($\downarrow$). Depth denotes effective depth. Darker/lighter blue marks the best/second-best results.}
\label{tab:olmo-transfer}
\centering
\papertablestyle
\begin{tabular}{lrrr}
\toprule
Model & Depth & CE $\downarrow$ & PPL $\downarrow$ \\
\midrule
Dense & 8 & 3.508664 & 33.403615 \\
Full Loop ($K=2$) & 16 & \secondresult{3.501389} & \secondresult{33.161466} \\
AsymLoop ($K=3$) & 16 & \bestresult{3.341073} & \bestresult{28.249416} \\
\bottomrule
\end{tabular}
\end{table}

\paragraph{Results.}
Table~\ref{tab:olmo-transfer} shows that AsymLoop achieves the lowest validation cross-entropy and perplexity among all three models. Compared with the depth-matched Full Loop, AsymLoop reduces perplexity by 14.8\%. These results demonstrate that its language-modeling benefits extend to an OLMo-style backbone, supporting its applicability across backbone architectures.

\section{Supervised Fine-Tuning on GSM8K}
\label{app:gsm8k-sft}

To evaluate mathematical reasoning after supervised fine-tuning, we fine-tune the 1B pretrained models on GSM8K~\citep{cobbe2021gsm8k}. We compare Dense, Full Loop, and AsymLoop after three epochs of full-parameter adaptation.

\subsection{Results after Three SFT Epochs}
\label{app:sft-final}

\begin{table}[H]
\caption{\textbf{GSM8K accuracy of 1B models after three epochs of full-parameter supervised fine-tuning.} The best result is highlighted in blue.}
\label{tab:sft-comparison}
\centering
\papertablestyle
\begin{tabular}{lrr}
\toprule
Model & GSM8K Accuracy (\%) $\uparrow$ & Relative gain over Dense (\%) $\uparrow$ \\
\midrule
Dense & 8.94 & -- \\
Full Loop ($K=2$) & 10.24 & +14.54 \\
AsymLoop ($K=3$) & \bestresult{11.75} & \bestresult{+31.43} \\
\bottomrule
\end{tabular}
\end{table}

Table~\ref{tab:sft-comparison} shows that AsymLoop achieves the highest GSM8K accuracy among the three models. Under matched fine-tuning schedules, AsymLoop reaches 11.75\% accuracy, improving over Full Loop by 1.52 percentage points. These results suggest that AsymLoop's pre-training benefits also extend to mathematical reasoning after supervised fine-tuning.

\subsection{Training and Evaluation Protocols}
\label{app:sft-protocol}

\paragraph{Models and optimization.}
We fine-tune the 1B pretrained models using AdamW with a global batch size of 32, a peak learning rate of $10^{-4}$, and cosine decay with 10\% warmup. We report results after three epochs using a single random seed. 

\paragraph{Data and supervision.}
We use 6,973 GSM8K training examples for supervised fine-tuning and reserve 500 for development, with a context length of 1024. Training supervises the worked solutions and final answers, excluding question and padding tokens from the loss.

\paragraph{Evaluation.}
We evaluate all 1,319 test questions using greedy decoding without demonstrations, allowing up to 512 generated tokens. Accuracy is computed by comparing the normalized final numerical answer with the reference.

\section{Sparse Mixture-of-Experts Models}
\label{app:moe}

We investigate whether AsymLoop's benefits extend to sparse mixture-of-experts models~\citep{muennighoff2025olmoe}. Using the 60M and 135M backbone configurations, we compare MoE variants of Dense, Full Loop, and AsymLoop, where Dense denotes the non-recurrent architecture.

\subsection{Experimental Setup}

We replace each feed-forward layer with eight SwiGLU experts and top-2 routing, without token dropping. Each expert has intermediate dimension 1024, preserving the active expert-matrix parameter count of the original dense FFN. Full Loop repeats the backbone twice, while AsymLoop divides it equally into an encoder and a decoder and repeats the decoder three times, matching the effective depth of Full Loop.

Experts and routers are shared across recurrences, with routing decisions recomputed at each iteration. Training combines cross-entropy with load-balancing and router z-loss terms weighted by 0.01 and 0.001, respectively. Validation cross-entropy excludes these auxiliary losses.

We train all models on FineWeb-Edu-dedup using AdamW with a cosine learning-rate schedule, sequence length 1024, and global batch size 512. The 60M and 135M configurations use 1.6B and 3B training tokens, respectively. We report validation cross-entropy and perplexity on a shared held-out set.

\subsection{Validation Results}

\begin{table}[H]
\caption{\textbf{Validation performance of sparse MoE models.} Cross-entropy and perplexity ($\downarrow$) with eight experts and top-2 routing. Darker/lighter blue marks the best/second-best results at each model scale.}
\label{tab:moe-main}
\centering
\papertablestyle
\begin{tabular}{lrrr}
\toprule
Model & Effective depth & CE $\downarrow$ & PPL $\downarrow$ \\
\midrule
\multicolumn{4}{l}{\textit{60M configuration: 1.6B training tokens}} \\[2pt]
Dense + MoE & 6 & 2.9856 & 19.798 \\
Full Loop K2 + MoE & 12 & \secondresult{2.9363} & \secondresult{18.845} \\
\textbf{AsymLoop K3 + MoE} & 12 & \bestresult{2.8986} & \bestresult{18.149} \\
\midrule
\multicolumn{4}{l}{\textit{135M configuration: 3B training tokens}} \\[2pt]
Dense + MoE & 12 & 2.7489 & 15.625 \\
Full Loop K2 + MoE & 24 & \secondresult{2.7154} & \secondresult{15.111} \\
\textbf{AsymLoop K3 + MoE} & 24 & \bestresult{2.6690} & \bestresult{14.426} \\
\bottomrule
\end{tabular}
\end{table}

Table~\ref{tab:moe-main} shows that AsymLoop achieves the lowest validation cross-entropy and perplexity at both model scales. At 135M, it reduces perplexity by 4.5\% relative to the depth-matched Full Loop. These results show that AsymLoop remains effective with sparse experts, extending its applicability beyond dense feed-forward layers.

\end{document}